\documentclass[11pt]{article}
\usepackage{acl2014}
\usepackage{times}
\usepackage{url}
\usepackage{latexsym}
\usepackage[utf8]{inputenc}
\usepackage{graphicx}

\usepackage{array}

\makeatletter
\def\barecite{\def\citename##1{{\frenchspacing##1,} }\@internalcitec}
\def\@internalcitec{\@ifnextchar [{\@tempswatrue\@citexc}{\@tempswafalse\@citexc[]}}
\def\@citexc[#1]#2{\if@filesw\immediate\write\@auxout{\string\citation{#2}}\fi
  \def\@citea{}\@barecite{\@for\@citeb:=#2\do
    {\@citea\def\@citea{;\penalty\@m\ }\@ifundefined
       {b@\@citeb}{{\bf ?}\@warning
       {Citation `\@citeb' on page \thepage \space undefined}}%
{\csname b@\@citeb\endcsname}}}{#1}}
\def\@barecite#1#2{{#1\if@tempswa, #2\fi}}
\makeatother

\title{Readability-based Sentence Ranking for\\ Evaluating Text Simplification}

 \author{Sowmya Vajjala \\
Iowa State University, USA \\
  {\tt sowmya@iastate.edu} \\\And
  Detmar Meurers \\
University of Tuebingen, Germany \\
  {\tt dm@sfs.uni-tuebingen.de} \\
 }

\begin{document}
\maketitle
\begin{abstract}
  We propose a new method for evaluating the readability of simplified
  sentences through pair-wise ranking. The validity of the method is
  established through in-corpus and cross-corpus evaluation
  experiments. The approach correctly identifies the ranking of
  simplified and unsimplified sentences in terms of their reading
  level with an accuracy of over 80\%, significantly outperforming
  previous results.
  
  To gain qualitative insights into the nature of simplification at
  the sentence level, we studied the impact of specific linguistic
  features. We empirically confirm that both word-level and syntactic
  features play a role in comparing the degree of simplification of
  authentic data.

  To carry out this research, we created a new sentence-aligned corpus
  from professionally simplified news articles. The new corpus
  resource enriches the empirical basis of sentence-level
  simplification research, which so far relied on a single
  resource. Most importantly, it facilitates cross-corpus evaluation
  for simplification, a key step towards generalizable results.
\end{abstract}

\section{Introduction}
\label{sec:intro}

Text simplification is the process of simplifying the linguistic form
of a text without losing its meaning. It has applications in several
domains ranging from language learning \cite{Petersen.Ostendorf-07}
and biomedical information extraction \cite{Jonnalagadda.Gonzalez-10}
for human readers all the way to automatic simplification designed to
improve parsing by machines \cite{Chandrasekar.Doran.ea-96}.  While
manual simplification relies entirely on expert writers,
semi-automatic approaches serve as an assistive tool for writers,
alerting them of text passages that may be difficult to read for the
target audience and indicating how to rewrite them
\cite{Candido.Maziero.ea-09}. Automatic text simplification
approaches, generating simplified text from an unsimplified version by
means of hand-crafted rules, data-driven methods, or hybrid
techniques have also been proposed (e.g.,
\barecite{Woodsend.Lapata-11,Siddharthan.Mandya-14}).

The nature of the simplification performed depends on the purpose of
the approach. Thus, the evaluation of a system that aims to improve the parser speed 
\cite{Chandrasekar.Doran.ea-96} also differs from one that was developed to support spoken language understanding \cite{Tur.Hakkani-Tuer.ea-11}In an educational context, typically the purpose is to
adapt the text to a level of complexity facilitating comprehension by
the target audience, such as language learners or students at a
particular grade level. It thus is important to
be able to evaluate the complexity of simplified and unsimplified
versions of a text -- which is the issue we address in this paper. The
approach is equally applicable to identifying those parts of a text
that are particularly complex and thus constitute good targets for
simplification.

Text simplification is generally evaluated in one of three ways:
through small-scale user evaluations, with a machine translation
metric, or using readability measures \cite{Siddharthan-14}.  We
explore the third option. Evaluating text simplification using
readability assessment is typically carried out with traditional
readability formulae. For example, \newcite{Woodsend.Lapata-11} make
use of the Flesch-Kincaid Reading Ease formula. Such readability
formulae are based on surface-level features: the average sentence
length, word length, or lists of difficult words
(cf.~\barecite{DuBay-06}). While such features often correlate with
the actual causes of difficulty in a piece of text (e.g., complex
syntax, infrequent words), manipulating these surface features does
not necessarily result in more readable texts
(cf.~\barecite{Klare-74}); they merely are surface indicators of a
broad range of underlying linguistic and psychological characteristics
of authentic texts targeting different audiences.  Recent research
also showed that more sophisticated, linguistically-grounded models
support a more reliable assessment of readability (e.g.,
\barecite{Nelson.Perfetti.ea-12}).  Although readability assessment is
primarily studied at a text-level, recent research explored it for
sentences as well (e.g.,
\barecite{Napoles.Dredze-10,Medero.Ostendorf-11,Pilan.Volodina.ea-14,DellOrletta.Wieling.ea-14,Vajjala.Meurers-14}).
Being able to assess the readability at the sentence level is
important to identify targets for simplification and to quantify the
degree of simplification performed for a given sentence.

\newcite{Vajjala.Meurers-14} studied sentential readability for text
simplification. They show that a relative comparison instead of an
absolute classification is better suited to identifying the difficult
sentences compared to simplified versions. They used a regression
model trained on whole documents to compare the readability of
parallel sentences from the sentence-aligned Wiki-SimpleWiki corpus
\cite{Zhu.Bernhard.ea-10}. While agreeing with the general
perspective, we propose a ranking approach which more directly
captures the idea of relative levels of readability, and we show that
it significantly outperforms the state of the art for evaluating
sentential simplification.

To support an evaluation of sentential simplification across different
corpus resources, testing whether something general has been learned,
we created a new corpus of aligned simple-complex sentence pairs. We
collected the data from the web site \mbox{OneStopEnglish.com}, which
offers manually simplified versions of news articles for language
learners. We tested our approach with this corpus, with the standard
Wikipedia-SimpleWikipedia sentence-aligned corpus, and using
cross-corpus evaluation. We show that our approach outperforms
previous approaches for identifying sentential simplifications and
that the performance generalizes across corpora. 

In terms of a qualitative analysis, we compare groups of features in
terms of their contribution to the ranking model and find that both
word-level and sentence-level properties play a role in ranking the
sentences by their reading level. While the psycholinguistic measures
of word properties figure prominently among the top features, the
best-performing model consists of all the features.

In sum, the contributions of this paper are:
\begin{enumerate}
\item We propose an approach to evaluate text simplification methods
  in terms of reading levels. Through multiple cross-corpus tests, we
  show that the approach performs at an accuracy of over 80\%.
\item We compiled a new corpus of sentence-level simplifications to
  obtain a broader empirical basis on which to evaluate and train text
  simplification systems.
\item We explored the role of individual features and feature groups
  for this task, including a comparison them across corpora.
\item In terms of the practical application context we are targeting,
  the quantitative and qualitative results in this paper establish
  that the approach can meaningfully be used in practice to evaluate
  simplification systems developed with the aim of reducing the
  difficulty of informative text for language learners.
\end{enumerate}

The paper is organized as follows: Sections~\ref{sec:corpora} and
\ref{sec:setup} describe the corpora and feature set we used.
Section~\ref{sec:expts} discusses our experiments and their
results. Sections~\ref{sec:related} and \ref{sec:conclusion} put our
research in context and conclude the paper.

\section{Corpora}
\label{sec:corpora}

The practical purpose of our approach is to evaluate text
simplification approaches aimed at helping language learners. Hence we
train and test our models on two corpora created with this target
audience in mind.

\subsection{OneStopEnglish corpus}
OneStopEnglish (OSE) is a resource website for English teachers
published by the Macmillan Education Group. They offer Weekly News
Lessons\footnote{\url{http://www.onestopenglish.com/skills/news-lessons/weekly-topical-news-lessons}}
consisting of news articles sourced from the newspaper \textit{The
  Guardian}. The articles are rewritten by teaching experts in a way
targeting English language learners at three reading levels
(elementary, intermediate, advanced). We obtained permission 
to use the articles for research purposes and downloaded the weekly
lessons from September 2012--March 2014, which resulted in a
collection of 76 article triplets. Each article is included in its
elementary, its intermediate, and its advanced version so that overall
the corpus contains 228 articles.

\paragraph{Corpus pre-processing} The weekly lessons are pdf files
consisting of a pre-test about the topic of the article, the
re-written news article, and exercises related to the article. We
first parsed the pdfs using
iTextPDF\footnote{\url{http://itextpdf.com}} to extract the article
text, excluding everything else. Since our aim is to compare different
versions of a sentence, we took each article triplet and
sentence-aligned two at a time using TF-IDF and cosine similarity,
following previous research on monolingual sentence alignment
\cite{Nelken.Shieber-06,Zhu.Bernhard.ea-10}.

\subparagraph{OSE3} For the sentences which exist in all three
versions of an article (elementary, intermediate, advanced), we obtain
a triplet of sentences.  We selected all triplets for which each pair
of sentences differed and was above a minimum similarity threshold of
0.7 (based on manual qualitative analysis using different
thresholds). Overall, we identified 837 sentence triplets and refer to
this corpus as OSE3.\footnote{We will share our sentence-aligned
  corpora for research purposes under a standard CC BY NC SA license.}
An example of a sentence rewritten across the three levels is shown
below:

\begin{description}
\item[Adv:] \textit{In Beijing, mourners and admirers made their way
    to lay flowers and light candles at the Apple Store.}
\item[Int:]\textit{In Beijing, mourners and admirers came to lay
    flowers and light candles at the Apple Store.}
\item[Ele:] \textit{In Beijing, people went to the Apple Store with
    flowers and candles.}
\end{description}

\subparagraph{OSE2} We compiled a second corpus consisting of pairs of
sentences, which we will refer to as OSE2. We extracted the 3113 pairs
of sentences (elementary-intermediate, intermediate-advanced, or
elementary-advanced) that differed and were above the minimum
similarity threshold.

\subsection{Wikipedia-SimpleWiki corpus}
Simple English Wikipedia (SimpleWiki) targets children and adults who
are learning
English,\footnote{\url{http://simple.wikipedia.org/wiki/Simple_English_Wikipedia}}
so a corpus of sentence pairs from Wikipedia and SimpleWiki
suits our goal to compare sentence-level text simplification. We use
the sentence-aligned corpus created by
\newcite{Zhu.Bernhard.ea-10}. They compiled a collection of $\sim$65k
parallel articles from Wikipedia and SimpleWiki to create a
sentence-aligned corpus consisting of $\sim$100k pairs. We used a
subset of this corpus consisting of 80,912 sentence pairs, after
removing the sentence pairs that are identical in both versions.

\section{Features and Methods}
\label{sec:setup}

\subsection{Features}
\newcite{Vajjala.Meurers-14} introduce a range of lexical, syntactic,
morphological, and psycholinguistic features to build a document-level
readability model that performed on par with existing commercial and
academic systems on the Common Core State Standard test set for
English \cite{ccsso-b-10}. They show that the model can also be
applied at the sentence-level.

Given our goal of studying the effectiveness of ranking over
regression and the relevance of specific features for sentence-level
ranking, we built our models using the same feature set they used,
which makes a direct comparison possible. The feature set of consists
of four groups of features:

\begin{itemize}
\item \textsc{Lex} consists of lexical diversity and density features
  primarily based on type-token and POS ratios, inspired by
  Second Language Acquisition (SLA) research, and lexical semantic
  properties from WordNet \cite{Miller-95} such as the average number
  of senses of a word.

\item \textsc{Syn} includes syntactic features based on specific
  patterns extracted from parse trees, including measures of syntactic
  complexity from SLA research.

\item \textsc{Cel} is a group of features encoding morpho-syntactic
  properties of lemmas, estimated using the Celex \cite{celex-95}
  database.

\item \textsc{Psy} contains word-level psycholinguistic features such
  as concreteness, meaningfulness and imageability extracted from the
  MRC psycholinguistic database \cite{Wilson-88} and various Age of
  Acquisition (AoA) measures released by
  \newcite{Kuperman.Stadthagen-Gonzalez.ea-12}.
\end{itemize}

\subsection{Methods}
\label{subsec:methods}

We model sentential complexity as a pair-wise ranking
problem. Pair-wise ranking is one of the \textit{learning-to-rank}
approaches, typically used in information retrieval for ranking search
results \cite{Li-14}. In that scenario, it is used to compare a pair
of documents in terms of their relevance to a given query. In our
case, the aim of the ranker given a pair of sentences (where one is
the simplified version of the other) is to predict which one of them
is simpler than the other. Thus, the learning problem for us is to
compare sentence pairs within a group of sentences and rank them based
on their complexity, trying to minimize inversion of ranks.

To apply ranking, we need to have a numeric score for (the feature
vector of) each sentence. In Wiki and OSE2, we assigned a reading level of 2 to
the more difficult version and 1 to its simplified version in the
sentence pair. For the sentence triplets in OSE3, we used the
sentences scores 3 (advanced), 2 (intermediate), and 1
(elementary). Since pair-wise ranking only considers relative ranks,
the ranking procedure is not dependent on the specific absolute
reading level of a sentence. In the case of sentences that were split into
two in the simplified version, we scored both the simple sentences as
1 so that no pair-wise constraints are generated between them.

\paragraph{Ranking:} We explored three ranking algorithms.

\subparagraph{RankSVM \cite{Herbrich.Graepel.ea-00}} 
transforms ranking into a pair-wise classification problem and uses a Support Vector Machine for learning to rank the sentence
pairs. It is one of the most commonly used ranking algorithms in NLP tasks. and was also employed in a related task, for ranking children's literature texts based on their reading level \cite{Ma.Fosler-Lussier.ea-12}.

\subparagraph{RankNet \cite{Burges.Shaked.ea-05}} is a pair-wise
ranking algorithm that is a modified version of a traditional
back-propagation-based neural network, applied to ranking problems. It
is known to perform well in practice and was successfully used in a
real-life search engine to rank search results.

\subparagraph{RankBoost \cite{Freund.Iyer.ea-03}} is an algorithm that
uses boosting for pair-wise ranking. It uses a linear combination of
several weak rankers to produce the final ranking. The algorithm was
typically used in collaborative filtering problems.

We used publicly available implementations of these algorithms for
training our models: SVM$^{rank}$
\cite{Joachims-06}\footnote{\url{http://www.cs.cornell.edu/people/tj/svm_light/svm_rank.html}}
for RankSVM and
RankLib\footnote{\url{http://sourceforge.net/p/lemur/wiki/RankLib}}
software for RankNet and RankBoost.

\paragraph{Evaluation:}
Since our learning goal is to minimize the number of wrongly ranked
pairs, we evaluate the approach in terms of the percentage of
correctly ordered pairs. In other words, we report the percentage of
pairs in which the difficult version gets a higher rank than its
simplified counterpart. We refer to this as accuracy.

\section{Experiments}
\label{sec:expts}

\subsection{Reference performance on Wiki dataset}

We start with an experiment directly comparing the ranking approach
with the results reported in \cite{Vajjala.Meurers-14} on the
Wiki-SimpleWiki data set. They used a regression model trained on
documents to get the reading levels for sentences. Their model
identified the rank order correctly in 59\% of the cases and assigned
equal score to both versions of the sentence in 11\% of the cases. So,
randomly considering half of the 11\% cases as correctly ordered and
half as wrongly ordered, one obtains a ranking accuracy of 64.5\% for
their model.

Replacing regression with ranking, we trained a model using
SVM$^{Rank}$ on the entire Wiki dataset in a 10-fold cross validation
(CV) setup. The ranking model achieves an accuracy of
82.7\%, which is a significant improvement over the baseline (p $<$ 0.01). The Standard Deviation between the ten folds is
8.4\%. This high level of variability suggests that the nature of
what constitutes simplifications in SimpleWiki varies significantly,
as may be expected for a collaborative editing setup -- a
potentially interesting issue to explore in the future.

As several text simplification approaches
\cite{Zhu.Bernhard.ea-10,Woodsend.Lapata-11} used the Flesch-Kincaid
Grade Level formula, which is based on the average word and sentence
length as surface features, as a readability measure for text
simplification, we also determined the accuracy of ranking the
sentences in the Wiki-SimpleWiki data using this formula and obtained
an accuracy of 72.3\%.

As summed up in Table~\ref{tab:first-experiment}, on the
Wiki-SimpleWiki dataset the ranking approach clearly outperforms the
regression approach of \newcite{Vajjala.Meurers-14} and the
Flesch-Kincaid readability formula.  The rich linguistic feature set
we have adapted from \newcite{Vajjala.Meurers-14} thus can clearly
outperform the surface-based readability formula, but the richer
information only becomes effective when the relative readability of
pairs of sentences is learned using a dedicated ranking algorithm.

\begin{table}[htb!]
  \centering
  \begin{tabular}{|l|c|}\hline
Approach & Accuracy\\\hline
\newcite{Vajjala.Meurers-14} & 64.5\%\\\hline
    Flesch-Kincaid formula &  72.3\%\\\hline
Our RankSVM approach & 82.7\%\\\hline
  \end{tabular}
  \caption{Ranking accuracy on Wiki-SimpleWiki}
  \label{tab:first-experiment}
\end{table}

To explore things further, we next compared different ranking
approaches and tested the generalizability of the ranking approach in
a cross-corpus setup and in multi-level simplification scenarios.

\subsection{Ranking algorithms and generalizability}
To compare the three ranking algorithms introduced in
Section~\ref{subsec:methods}, we first trained ranking models for the
\textsc{Wiki} and \textsc{OSE2} corpora. To make the results
comparable for these two corpora, we selected 2,000 sentence pairs for
each of the training sets (\textsc{Wiki-Train}, \textsc{OSE2-Train}),
and used the remaining part as the test set (\textsc{Wiki-Test:} 78,912
pairs, \textsc{OSE2-Test:} 1,113 pairs).

Table~\ref{tab:rankers1} presents the performance of the three ranking
algorithms on the two training sets for within and cross-corpus
evaluation. As a baseline, the Flesch-Kincaid formula results in
69.0\% for \textsc{Wiki-Test} and 69.6\% for \textsc{OSE2-Test}.


\begin{table}[htb!]
\resizebox{.48\textwidth}{!}{%
\begin{tabular}{|c|c||c|c|c|}\hline 
\textsc{Train} &  \textsc{Test} &   RankSVM &   RankNet &  RankBoost \\
        \hline 
\textsc{Wiki} & \textsc{Wiki} &\textbf{81.8\%}&72.5\%&76.4\% \\
        \hline 
\textsc{Wiki} & OSE2 &74.6\%&59.1\%&70.2\% \\
        \hline 
OSE2 & \textsc{Wiki} &77.5\%&73.8\%&74.8\% \\
        \hline 
OSE2 & OSE2 &\textbf{81.5\%}&69\%&75.5\% \\
        \hline
     \end{tabular}}
    \caption{Accuracies for the three rank algorithms}
    \label{tab:rankers1}
\end{table}


\noindent RankSVM performs best among the ranking algorithms we
tried. This also generalizes to the cross-corpus tests. In the
following experiments, we therefore only report the results for RankSVM.

Cross-corpus evaluation always shows a drop in performance. The drop
is smaller for the model trained on the OSE2 corpus, which suggests
that the OSE2 corpus covers a broader, more representative range of
simplifications. Taking that idea further, we explored improving
cross-corpus performance using two methods enriching the training
data.

\subsection{Improving cross-corpus performance}
First, we combined the two training sets to create a  hybrid training set
\textsc{Wiki-OSE2-Train}, which should increase the representativity
and range of the simplifications included in the training data.

Second, we used the three level corpus \textsc{OSE3} to train the
ranker to simultaneously consider a broader range of simplifications:
the ranker will learn a single set of weights for ranking the three
pairs in a set for OSE3, instead of three sets of weights for ranking
each pair independently. We randomly selected 750 sentence triplets
from the \textsc{OSE3} corpus as training set (\textsc{OSE3-Train}),
leaving the remaining 87 as held-out test set
(\textsc{OSE3-Test}). Table~\ref{tab:newttsets} shows the results on
the three test sets for the models trained on \textsc{Wiki-OSE2-Train}
and \textsc{OSE3-Train}. The baseline accuracy obtained using the
Flesch-Kincaid formula for \textsc{OSE3-Test} is 71.6\%.


\begin{table}[htb!]
\resizebox{.48\textwidth}{!}{%
    \begin{tabular}{|c|c|c|}
      \hline 
                               &\textsc{Wiki-OSE2-Train} & \textsc{OSE3-Train}\\
      \hline \textsc{Wiki-Test}&81.3\%&78.6\%\\
      \hline \textsc{OSE2-Test}&80.7\%&82.4\%\\
       \hline \textsc{OSE3-Test}&79.7\%&\phantom{$^9$}79.7\%\footnotemark\\
      \hline
    \end{tabular}}
    \caption{Accuracies for the extended training sets}
    \label{tab:newttsets}
\end{table}

\footnotetext{The identical overall performance of both models on the
  \textsc{OSE3-Test} set differs in terms of individual instances.}

\noindent As expected, the accuracy for the combined, more varied
training set \textsc{Wiki-OSE2-Train} results in a comparable
performance across the three tests sets. It seems to account for a
broader range of simplification options. The results for the
\textsc{OSE3-Train} set, providing the ranker with triples over
which to learn the weights, are less clear.

Overall, the fact that cross-corpus and same-corpus results are
relatively close together supports the assumption that reliable
sentence-level readability ranking models which generalize across very
different data sets can be built.

\subsection{Influence of the amount of training data}

\begin{figure}[htb!]
 \includegraphics[clip,trim=2cm 1.8cm 2.6cm 12.2cm,width=0.48\textwidth]{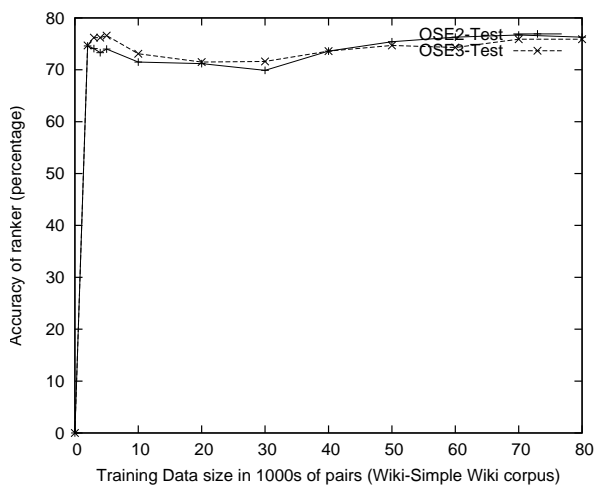}
  \caption{Accuracies for increasing training size}
  \label{fig:accbytrainingsize}
\end{figure}

While the \textsc{Wiki-OSE2-Train} contains more diverse data, it also
contains twice as much data as the two smaller training sets it
consists of. To isolate the effect of the training data size, we
explored how the low cross-corpus performance of 74.6\% of for the
\textsc{Wiki-Train} model on the \textsc{OSE2-Test} we saw in
Table~\ref{tab:rankers1} is improved simply by increasing the amount
of training data. We therefore trained on increasingly larger portions
of the Wiki-SimpleWiki data set up to the full set of 80k pairs. We
tested on \textsc{OSE2-Test} and additionally on \textsc{OSE3-Test} to
lower the risk of idiosyncrasies specific to a single test
set. Figure~\ref{fig:accbytrainingsize} 
shows the accuracies for models trained on increasing amount of
Wiki-SimpleWiki training data.

The curve is essentially flat, with the model on the largest training
set (80k) reaching an accuracy of 76.3\%, less than two percent above
the result we obtained using only 2k pairs for training, and
significantly below the 80.7\% obtained for the model trained on the
4k \textsc{Wiki-OSE2-Train} set (cf. Table~\ref{tab:newttsets}). The
Wiki-SimpleWiki data thus does not seem to offer the variety of
simplification needed to generalize better to the \textsc{OSE} test
sets.

\subsection{Feature Selection}

Turning from experiments establishing the overall validity of the
approach to the impact of the different features, the next experiments
explore feature selection. In addition to characterizing how much can
be achieved with how little, feature selection gives us an opportunity
to better understand the linguistic characteristics of
simplification. We explored which features contribute the most as
single features or as feature groups. 

\begin{figure}[htb!]
  \includegraphics[clip,trim=1.9cm 1.8cm 2.2cm 11.2cm,width=0.48\textwidth]{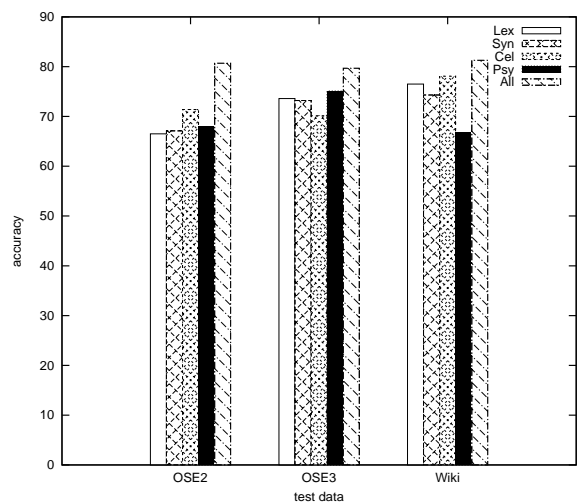}\vspace{-2.6mm}
  \caption{Performance of different feature groups}
  \label{fig:fgperformance}
\end{figure}

\paragraph{Impact of feature groups:} 
First, we investigated the contribution of different feature groups to
ranking accuracy. Figure~\ref{fig:fgperformance} presents the performance
of ranking models trained using the four feature groups (\textsc{Lex},
\textsc{Syn}, \textsc{Cel}, \textsc{Psy}) and a model trained with all
features. We train on the \textsc{Wiki-OSE2-Train} dataset, which
as we saw in Table~\ref{tab:newttsets} generalized well across
different test sets.

Figure~\ref{fig:fgperformance} shows that the performance of the
individual feature groups varies with the test-sets used. For example,
\textsc{Cel} features result in lower accuracy for \textsc{OSE3-Test}
compared to other feature groups, whereas \textsc{Psy} features
performed poorly for \textsc{Wiki}. For all test sets, the model
trained with all the features outperforms the individual feature group
models. For a generalizing approach to the evaluation of text
simplification, modeling multiple dimensions of readability thus is
more useful than choosing a single aspect.

\paragraph{Impact of individual features:} 
To understand the linguistic characteristics of simplification, it is
useful to identify which individual features are more informative for
these authentic data sets. Hence, we trained single feature ranking
models with each of the training sets and ranked the features based on
their performance on the test sets. The list of single features 
achieving over 60\% for in-corpus test-set evaluation are shown in
Table~\ref{tab:topwiki} 
\begin{table}[htb!]
\centering\begin{tabular}{|>{\raggedright}p{4cm}|c|c|}
\hline
\textbf{Feature} & \textbf{Group} & \textbf{Accur.}\\
\hline
num. subtrees (\textsc{subtrees}) & SYN&72.1\%\\ 
\hline corrected type-token ratio (\textsc{cttr}) & LEX&70.4\%\\ 
\hline sentence length (\textsc{senlen}) & SYN & 69.7\%\\ 
\hline avg. Age of Acquisition Kup.-Lem. (\textsc{AoA}) & PSY & 64.8\%\\ 
\hline num. constituents per tree (\textsc{const}) & SYN&63.3\%\\ 
\hline avg. length of t-unit (\textsc{mlt}) &SYN&63.2\%\\ 
\hline
\end{tabular}
\caption{Performance of single feature ranking models for
  \textsc{Wiki-Train}/\textsc{Wiki-Test}}
\label{tab:topwiki}
\end{table}
for the \textsc{Wiki} data and
Table~\ref{tab:topose2} for the \textsc{OSE2} data.\footnote{We
  experimented with a range of AoA norms and lexical diversity
  measures, but for space reasons include only the best AoA and
  lexical diversity feature here. Interestingly, the accuracies
  obtained for the various AoA norms differed substantially, between
  37\% and 72.8\%, also due to coverage.}

\begin{table}[htb!]
  \centering
\begin{tabular}{|>{\raggedright}p{4cm}|c|c|}
\hline
\textbf{Feature} & \textbf{Group} & \textbf{Accur.}\\
\hline \textsc{AoA} & PSY& 72.8\%\\ 
\hline \textsc{cttr} & LEX& 66.7\%\\ 
\hline \textsc{subtrees} & SYN& 64.4\%\\ 
\hline avg. length of clause (\textsc{mlc}) &SYN &63.2\%\\ 
\hline avg. word imagery rating (\textsc{imagery}) &PSY& 63.2\%\\ 
\hline avg. word familiarity rating (\textsc{familiarity}) & PSY& 63.2\%\\ 
\hline avg. Colorado meaningfulness rating of words (\textsc{meaningfulness}) & PSY& 63.2\%\\ 
\hline avg. concreteness rating (\textsc{concreteness}) &PSY&  61.7\%\\ 
\hline
\end{tabular}
\caption{Performance of single feature ranking models for
  \textsc{OSE2-Train}/\textsc{OSE2-Test}}
\label{tab:topose2}
\end{table}

For the \textsc{Wiki} data only six features individually performed
with an accuracy above 60\%: four \textsc{Syn}, one \textsc{Lex}, one
\textsc{Psy}. For the \textsc{OSE2} data, this was the case for eight
features: two \textsc{Syn}, one \textsc{Lex}, five \textsc{Psy}.
Both lists thus include a combination of word-level and syntactic
features, with syntactic simplification playing more of a role for the
\textsc{Wiki} dataset and lexical choices relating to psycholinguistic
characteristics being more relevant for the \textsc{OSE2}
data. 

Sentence length is more predictive for \textsc{Wiki} than for
\textsc{OSE2}, probably because the \textsc{Wiki} dataset contains a
lot of deletions ($\sim$45\% of the sentence pairs show major
deletions) compared to the \textsc{OSE} dataset, where sentences were
mostly rewritten or paraphrased instead of deleting content. In line
with this observation, sentence length as a single feature for
\textsc{OSE2-Test} data achieves an accuracy of only 57.5\%, compared
to the 69.7\% for \textsc{Wiki-Test}.

The role and interdependence of the different psycholinguistically
motivated features (age of acquisition, concreteness, meaningfulness,
imagery) for the \textsc{OSE2} data is interesting and would merit
further study. A good understanding of the role of these features
would be directly relevant for improving lexical simplification
approaches such as that of \newcite{Jauhar.Specia-12}, which already
integrates some features from the MRC psycholinguistic database to
rank word substitutes for lexical simplification.

\subsection{Simplification at different levels}

We next explored, whether the nature of the simplification differs
between advanced sentences being simplified compared to intermediate
sentences being (further) simplified. To investigate this, we split
the \textsc{OSE3-Train} and \textsc{OSE3-Test} datasets into two pairs
of datasets \textsc{Adv-Int-Train, Adv-Int-Test} and
\textsc{Int-Ele-Train, Intr-Ele-Test}. Table~\ref{tab:simplevels}
shows the differences in the performance of the ranking approach
between the two levels of simplification.

\begin{table}[htb!]
\centering\begin{tabular}{|l|c|c|}
\hline & \multicolumn{2}{c|}{Training Data}\\
& \textsc{Adv-Int}&\textsc{Int-Ele}\\
\hline \textsc{Adv-Int-Test} &73.6\%&74.7\%\\
\hline \textsc{Int-Ele-Test}&81.6\%&80.5\%\\
\hline
\end{tabular}
\caption{Simplification at different levels}
\label{tab:simplevels}
\end{table}

The performance on the \textsc{Int-Ele-Test} set is better than that
on the \textsc{Adv-Ele-Test} set, independent of whether the model was
trained on the \textsc{Adv-Int} or \textsc{Int-Ele} training data. To
understand the reason, we explored the nature of the simplification
involved at these two different levels by testing the predictive power
of individual features.

Table~\ref{tab:topintele} shows the list of features that individually
achieved an accuracy of over 60\% for intermediate to beginner level
simplification. For advanced to intermediate, only AoA features
achieved an accuracy of above
60\%. 
\begin{table}[htb!]
\centering\begin{tabular}{|l|l|l|}
\hline
\textbf{Feature} & \textbf{Group} & \textbf{Accur.}\\
\hline
\textsc{AoA}&\textsc{Psy}&77\%\\
\textsc{imagery} &\textsc{Psy}&67.8\%\\
\textsc{cttr}&\textsc{Lex}&67.8\%\\ 
\textsc{meaningfulness} & \textsc{Psy}&66.7\%\\
\textsc{concreteness}&\textsc{Psy}&65.5\%\\
\textsc{familiarity}&\textsc{Psy}&64.4\%\\
\textsc{mlc}&\textsc{Syn}&64.4\%\\
\textsc{subtrees}&\textsc{Syn}&64.4\%\\
avg. senses per word&\textsc{Lex}&64.4\%\\
\hline
\end{tabular}
\caption{Accuracy of single feature ranking models for
  \textsc{Int-Ele} simplification}
\label{tab:topintele}
\end{table}

The better overall performance at the intermediate to elementary
simplification level and the higher number of informative features at
that level indicate that the nature of the simplification between
advanced and intermediate sentences is more subtle -- and possibly the
already broad feature set warrants further extension to capture
additional characteristics of more advanced levels. 
%
%

For example, many of the syntactic features mentioned in the feature
selection discussion (\textsc{subtrees}, \textsc{mlc}, \textsc{mlt})
are correlated with text length. However, simplification can also
involve sentence rewrites that do not affect the sentence length as
such (e.g., paraphrasing, active/passive, reordering), which may
warrant inclusion of features targeting more specific syntactic
constructions or idiomatic word usage characteristic of advanced
levels of English.

\subsection{Error Analysis}
Finally, to understand if there is a systematic pattern in the errors
made by the ranker, we manually performed a qualitative analysis of
errors. For this, we took the results of training with
\textsc{OSE3-Train} data and testing with the \textsc{OSE3-Test}. This
is the smallest test set (87 triplets), which given the 79.7\%
accuracy allows us to analyze 53 misclassified pairs. The following
are four example sentence pairs/triplets from the test set. While the
first two were ranked correctly by the ranker, the last two illustrate
cases where the ranker failed.

\paragraph{Example 1}
  \begin{description}
\item[Adv:]  \textit{He warned that it was too early to use oxytocin
      as a treatment for the social difficulties caused by autism and
      cautioned against buying oxytocin from suppliers online.}
\item[Int:]  \textit{He warned that it was too early to use oxytocin
      as a treatment for the social difficulties caused by autism and
      said people should not buy oxytocin online.}
\item[Ele:]  \textit{He said that it was too early to use oxytocin as
      a treatment for the social difficulties caused by autism and
      said people should not buy oxytocin online.}
  \end{description}
  
\paragraph{Example 2}
  \begin{description}
\item[Int:] \textit{DNA taken from the wisdom tooth of a European
      hunter-gatherer has given scientists a glimpse of modern humans
      before the rise of farming.}
\item[Ele:] \textit{Scientists have taken DNA from the tooth of a
      European hunter-gatherer and have found out what modern humans
      looked like before they started farming.}
  \end{description}

\paragraph{Example 3}
  \begin{description}
  \item[Adv:] \textit{Its inventor, Bob Propst, said in 1997, ``the
      cubiclizing of people in modern corporations is monolithic
      insanity.''}
  \item[Int:] \textit{Its inventor, Bob Propst, said, in 1997, ``the use
      of cubicles in modern corporations is crazy.''}
  \item[Ele:] \textit{The inventor, Bob Propst, said, in 1997, ``the use
      of cubicles in modern companies is crazy.''}
  \end{description}
  
\paragraph{Example 4}
  \begin{description}
  \item[Adv:] \textit{A special ``auditor'' declares him 96.9\% ``made in
      France'' and Montebourg visits to present him with a medal.}
  \item[Int:] \textit{A special ``auditor'' declares him 96.9\% ``made in
      France'' and Montebourg visited to present him with a medal.}
\end{description}

In Example 1, the transformation from \textit{Adv} to \textit{Int} is
primarily paraphrasing (``and cautioned against buying oxytocin''
vs.~``and said people should not buy oxytocin'') where was the
transformation from \textit{Int} to \textit{Ele} is that of a simple
lexical substitution (``He warned'' vs.~``He said''). However, in
Example 2, there was more re-ordering and paraphrasing (``before the
rise of farming'' vs.~``before they started farming''). In both these
cases, our model correctly identified the changes as a
simplification. The model thus effectively identifies paraphrases and
lexical substitutions at multiple levels.

However, the model is not as effective for the sentence triplet in
Example 3. It provides the correct pairwise rankings \textit{Int} $>$
\textit{Ele} and \textit{Adv} $>$ \textit{Ele}, but then incorrectly
determines \textit{Int} $>$ \textit{Adv}.  So the model correctly
identified a simple lexical substitution between \textit{Int} and
\textit{Ele}, but failed to identify the transformation from ``the
cubiclizing of people'' to ``the use of cubicles'' and from
``monolithic insanity'' to ``crazy'' as a simplification. This could
possibly be because the parse structure as such did not alter much
despite the rephrasing and neither ``cubiclizing'' nor the noun or
lemmatized verb ``cubiclize'' exist in the psycholinguistic databases
we used. Including broad coverage frequency measures of word usage
could help address examples of this type. If the example at hand is
typical, for the purpose of simplification evaluation at issue here
word form frequencies would be preferable over lemma frequencies.

Finally, in Example 4 the only change between the two versions is a
tense difference (``visits'' vs.~``visited''), which the model fails
to rank correctly. It is debatable whether this change in tense indeed
represents a simplification so that the case does not provide useful
information on how to improve the approach.

Apart from the relevance of integrating broad-coverage frequency
measures characterizing word form usage, our qualitative error
analysis did not identify systematic failures of our models. The broad
coverage of linguistic features integrated in a ranking approach
successfully capture the relative differences in readability which
characterize authentic simplification data at the sentence level.

\section{Related Work}
\label{sec:related}

Evaluation of a text simplification approach is typically done in two
ways. Most approaches are evaluated by comparing sentences using a
combination of traditional readability measures, human fluency and
grammaticality judgments of the generated output, and machine
translation metrics (e.g.,
\barecite{Barlacchi.Tonelli-13,Stajner.Mitkov.ea-14,Siddharthan.Mandya-14}). Some
approaches evaluate the effect of text simplification on their target
audience in terms of human recall and comprehension (e.g.,
\barecite{Canning.Tait.ea-00,Williams.Reiter-08,Bradley-12}). Other
recent work reported the usage of linguistic complexity measures that
go beyond traditional readability formulae for evaluation of text
simplification at a document level \cite{Stajner.Saggion-13a}.

Comparing simplified versions of individual sentences with
unsimplified versions in terms of text complexity is a rather recent
endeavor. For example, sentence-level text complexity was explored in
Intelligent Computer Assisted Language Learning to identify suitable
sentences for creating learning exercises for German and Swedish
learners
\cite{Segler-07,Pilan.Volodina.ea-14}. \newcite{DellOrletta.Wieling.ea-14}
explored the linguistic nature of features contributing to sentential
readability in the context of developing Italian text simplification
system for adults with intellectual disabilities. However, the corpus
used does not provide parallel texts with easy and difficult versions.
In the absence of a sentence-aligned simplification corpus, the
authors treat each sentence in the easy-to-read texts as easy. As
\newcite[Fig.\,1]{Vajjala.Meurers-14b} showcases, this is a very
problematic assumption. Even for a sentence-aligned simplification
corpus such as the Wiki-SimpleWiki data set the only thing guaranteed
is that there is one sentence which is harder than the simple one.

\newcite{Napoles.Dredze-10} considered a binary classification of
Wiki-SimpleWiki at both text and sentence level, using a range of
lexical and syntactic features. They also work with the simplifying
assumption that all sentences in Wikipedia are difficult and those in
SimpleWiki are simple. An interesting aspect of the results of
\newcite{Napoles.Dredze-10} and also of
\newcite{DellOrletta.Wieling.ea-14} is that they achieve the highest
classification accuracy at text and sentence level when combining all
features.

\newcite{Vajjala.Meurers-14} compared sentences in terms of their
reading levels using their readability model and showed that
sophisticated linguistically motivated readability models can
effectively identify the differences between sentences. In the current
paper, we extend their research by exploring sentential simplification
evaluation as a ranking problem and showed that ranking achieves
superior performance to regression for this task.

Ranking has been used for readability assessment at the document level
\cite{Ma.Fosler-Lussier.ea-12} and for related tasks such as ordering
MT system output sentences in terms of their language quality
\cite{Avramidis-13}, and for ranking sentences in opinion
summarization \cite{Kim.Castellanos.ea-13}. To the best of our
knowledge, simplification evaluation was not explored as a ranking
problem before.







\section{Conclusions}
\label{sec:conclusion}

We presented an approach to evaluate automatic text simplification
systems in terms of the reading level of individual sentences. The
approach involves the use of a pairwise ranking approach to compare
unsimplified and simplified versions of sentences in terms of the
reading level. It identifies the order in terms of their reading level
correctly with an accuracy of over 80\%, the best accuracy for this
task we are aware of. We performed in-corpus and cross-corpus
evaluations with two very different sentence-aligned corpora and
showed that the approach generalizes well across corpora. The approach
performs at a level that should make it useful in practice to
automatically evaluate text simplification for language learners in
real-life educational settings. 

In terms of the linguistic nature of simplification, we studied the
role of individual features and groups of features in predicting the
ranking order between simplified and unsimplified versions of the
sentences. We found that for the OSE data set, psycholinguistic
features such as Age-of-Acquisition are the most predictive individual
features. For the Wiki-SimpleWiki data set, syntactic features also
figure prominently. However, an approach using the full range of
features systematically results in the best performance and
generalizes best in the cross-corpus settings.

Pursuing the question whether there is a singular notion of
simplification, we studied the differences in text simplification
occurring at different levels of readability. Our features identify
simplification between intermediate and elementary levels better
compared than between advanced and intermediate level. It is possible
that this is due to a higher degree of simplification between the
former, but we also plan to study whether other types of features
could be added to identify characteristics at higher levels of
readability, such as features targeting specific constructions or
idiomatic usage. The small qualitative error analysis we performed
revealed that a broad coverage method capturing the frequency of word
usage may further improve results.

To carry out the research, we created a new corpus of sentence-aligned
simplified texts based on OneStopEnglish texts rewritten by experts
for language learners into three reading levels. The new corpus
resource can empirically enrich future research on sentence-level
simplification, helping to ensure that the results obtained are more
generally valid than for the single Wiki-SimpleWiki resource that was
available so far.

\paragraph{Outlook}

Adding frequency features capturing word usage
and exploring feature selection in more detail by selecting
the best features for the ranker while removing the correlated ones
\cite{Geng.Liu.ea-07} are the immediate directions we would like to
pursue.

It would also be interesting to apply the approach to evaluate the
output of automatic text simplification systems and compare their
performance in terms of readability. Going beyond complexity, in the
long term it could be interesting to extend the approach to a full
framework for evaluating automatic text simplification systems by
integrating aspects of fluency and grammaticality.



\bibliographystyle{acl} 
\bibliography{icall}

\end{document}